\documentclass[11pt,a4paper]{article}
\usepackage{acl2021}
\usepackage{times}
\usepackage{latexsym}
\usepackage{microtype}
\usepackage{graphicx}
\usepackage{hyperref}
\usepackage{geometry}
\usepackage{makecell}
\usepackage{float}

\title{Multi-modal application: Image Memes Generation}

\author{Zhiyuan Liu, Chuanzheng Sun, Yuxin Jiang, Shiqi Jiang, Mei Ming
    \\
    \\ University of Southern California
    \\ {\tt \{zliu5480,chuanzhe,yjiang30,shiqijia,mming\}@usc.edu}}

\date{11/29/2021}

\begin{document}

\maketitle

\section{Introduction}
Meme is an interesting word. Internet memes offer unique insights into the changes in our perception of the world, the media and our own lives. If you surf the Internet for long enough, you will see it somewhere on the Internet. With the rise of social media platforms and convenient image dissemination, Image Meme has gained fame. Image memes have become a kind of pop culture and they play an important role in communication over social media, blogs, and open messages.

The word Meme is derived from the Greek word "Mimema" and refers to the idea being imitated. Richard Dawkins defined the word "meme in his book \textit{The Selfish Gene} \cite{dawkins2017selfish} published in 1976, which ascribed meme to an idea, behaviour or style that spreads from person to person within a culture.

Because memes are also often used, the demand for new memes is very strong. However, new memes need to be personally created by adding copywriting and synthesizing pictures. Since the pictures or captions used to create new memes will not always fit, not all memes can be perfectly transmitted to the author. Therefore the memes that users commonly use are those that already exist.

With the development of artificial intelligence and the widespread use of deep learning, Natural Language Processing (NLP) and Computer Vision (CV) can also be used to solve more problems in life, including meme generation. An Internet meme commonly takes the form of an image and is created by combining a meme template (image) and a caption (natural language sentence). In our project, we propose an end-to-end encoder-decoder architecture meme generator. For a given input sentence, we use the Meme template selection model to determine the emotion it expresses and select the image template. Then generate captions and memes through to the meme caption generator. Code and models are available at github \footnote{https://github.com/zliu5480/CSCI544Project}.

We summarize our achievements as follows:

\begin{itemize}
  \item We propose an end-to-end encoder-decoder architecture meme generator to produce an image meme for a given sentence.
  \item We train a text-emotion classifier using a BERT pre-trained model, which is called Meme Template Selection Model.
  \item We train a sequence to sequence networks with attention to convert input sentences to appropriate captions, which name is Meme Caption Generator.
\end{itemize}

\section{Related Work}
There are several kinds of research on caption generation and caption selection or meme generation. In 2005, Wang and Wen have proposed \cite{wang2015can} the approach based on the ranking algorithm to select captions from the existing corpus. According to Peirson et al. \cite{peirson2018dank}, they combined the Inception-v3 network and attention-based LSTM to produce the caption for memes. The method is an extension of Natural Language Description Generation.

Besides methods about selecting meme captions from corpus, researchers also proposed that the caption generation task can be aligned with the tasks like Sentiment Analysis, Neural Machine Translation, Image Caption Generation \cite{sadasivam2020memebot}. Socher et al. \cite{socher2013recursive} classified sentences based on the sentiments. Based on this idea, Huang et al \cite{huang2018automatic}. embedded sentiment representation into the Seq2Seq model to generate text with the desired sentiment. Miao et al. \cite{miao2019cgmh} used sampling techniques to generate sentences with desired emotions or keywords. Sutskever et al. \cite{sutskever2014sequence} and Vaswani et al. \cite{vaswani2017attention} came up with an encoder-decoder model to switch the sentence from source to the targeted one.

In addition to the researches started from sentences, Karpathy and Fei-Fei \cite{karpathy2015deep} designed the approach to encode the visual features of an image and use a decoder to generate a natural language description of the image.

For our approach which is combined by meme picture selection part and text summarization part. There are some related researches on these topics.
\paragraph{Picture Selection}
The main problem here is to understand the natural language. Bert \cite{devlin2018bert}, XLNet \cite{yang2019xlnet} and Roberta \cite{liu2019roberta} are transformer-based architectures which are widely used in Natural Language Understanding tasks. Moreover, for the meme selection, Sadasivam et al.\cite{sadasivam2020memebot} mentioned that pre-trained language representation models with a linear neural network can be used. 
\paragraph{Text Summarization}
For the meme caption generator, many authors choose to use a Long Short Term Memory (LSTM) or RNNs due to their well-established success on sequence tasks \cite{peirson2018dank}. According to Sadasivam et al.\cite{sadasivam2020memebot}, extracting the parts of speech of the input caption using a Part-Of-Speech Tagge and using the POS vector to learn from the existing sentences and to generalize caption is another type of method.

\section{Methods}
\subsection{Meme Template Selection Model}
For the Meme Template Selection Model, we made several text-emotion classifiers. We decided to analyze an input text’s emotion and choose the Meme Template. There are two datasets: one dataset (ISEAR) is produced by The International Survey on Emotion Antecedents and Reactions database \cite{scherer1994evidence}, which is a balanced data set and another dataset (dailydialog) is a manually labelled multi-turn dialogue dataset provided by Yanran Li \cite{li2017dailydialog}, which is not a balanced dataset. Considering the content of our project, we decided to use dailydialog dataset because it is more close to meme generation. We made a balanced dataset with 5 labels: joy, sad, anger, fear, and neutral. We tried several models to do the text emotion classification, BiLSTM model, Text-CNN model, DistilBERT, and our final fine tuned BERT model.

The evaluation of Meme template selection is just the evaluation of a normal text classifier, we use accuracy and f1 score to help select the best model.

\subsection{Meme Caption Generator }
Our understanding of the caption generation is that the model takes a sentence as input and it outputs a “meme-style” text that has the same meaning. Science the input and output are two different sentences with the same meaning, we could use sequence to sequence networks with attention.

Our initial thought is using Part-Of-Speech tagger to analyze the input text, so we can pass only useful text such as noun phrases and mask out other words. With Professor Ma’s suggestion, we realize that POS tags are not necessary for our approach. Instead, we should directly take input text as input.

Unfortunately, we cannot directly find dataset that can directly be used for caption generation, so we make our own data based on the meme template and meme captions we have. Our data has two parts: one is an input sentence and the other one is a suitable meme caption for the sentence. Due to the time and human resource we have, we only have 500 training data. Table \ref{Table 1} are some examples of our training data.

\begin{table*}[htbp]
	\centering
    	\begin{center}
    	\resizebox{\textwidth}{18mm}{
    		\begin{tabular}{|c|c|}
    			\hline Input & Output\\\hline
    			you can hug a cactus when you sad & sad? hug a cactus\\
                I need to lose weight, should I do heroin? & need to lose weight? do heroin\\
                you should wear outfit to keep yourself warm & i don't always wear outfit but when I do I keep myself warm\\
                i do not understand this, this is so complicated & wtf is this! i hate it so much\\
                I need money to buy an iphone & patient bear waits for his iphone\\
    			\hline
    		\end{tabular}}
    	\end{center}
	\caption{Input and Out}
	\label{Table 1}
\end{table*}

Science we created our own training data. The data is nice and clean. But we still applied preprocessing on the data: make data into lower case, separate characters such as: “,”, “.”, “!”, “?”, and remove extra spaces. Then we created a dictionary for word2index including special tokens like “\textless SOS\textgreater”, “\textless EOS\textgreater” and “\textless UNK\textgreater”.

\begin{figure}[H] 
\centering
\includegraphics[width=0.4\textwidth]{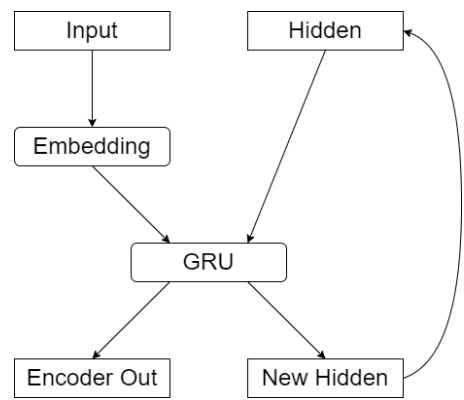} 
\caption{Encoder of S2S model}
\label{Fig1}
\end{figure}

In our encoder of sequence to sequence model, we use a simple embedding with gru to convert every word in the sentence into some value. For every input the encoder outputs an output and a hidden state for the next input word. The structure is shown in Figure \ref{Fig1}.

\begin{figure}[H] 
\centering
\includegraphics[width=0.4\textwidth]{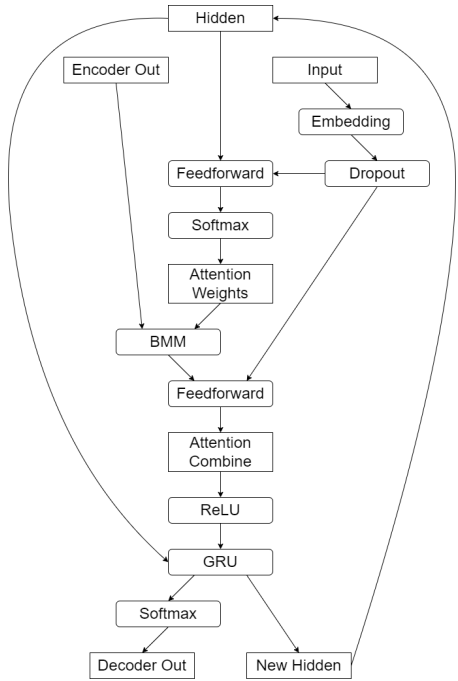} 
\caption{Decoder of S2S model}
\label{Fig2}
\end{figure}

In our decoder of sequence to sequence model we used attention mechanism and also gru to produce output and hidden. For the decoder there are three inputs: input text, encoder’s output and previous hidden. The first hidden is taken from the final hidden state of the encoder. To calculate the attention weights, we use a feed-forward layer with softmax. With attention weights, we apply matrix multiplication (BMM) with the decoder’s output and use another feedforward to get attention\_combine from the output of BMM and the embedding of the input. Lastly we put our attention\_combine into the gru and softmax to get our final output. The structure is shown in Figure \ref{Fig2}.

The evaluation of our Meme Generation is done by ourselves. The common evaluation methods for NLP tasks like BLUE do not really fit our project because we do not have a gold standard output. We make our own input text and let the model generate output to see if it meets the expectations. We evaluate each test input and output with three types: good, acceptable, bad.

\section{Experiments}
\subsection{Meme Template Selection Model}
For the Meme Template Selection Model, We try to build several models in order to pick the model with highest accuracy to implement in the following work.

First we built a BiLSTM model(\cite{huang2015bidirectional}), followed by two linear layers with 128 output nodes and 7 output nodes. After 20 epochs, the loss is converging and the best accuracy is 61\%.

The second model we trained is a Text-CNN model(\cite{zhang2015sensitivity}). We use the pre-trained GloVe word vector model to convert the word-embedding of each word, and convert and concatenate each sentence to obtain a two-dimensional matrix. Inside the CNN structure, we use one-dimensional convolution layers, design four different kernel sizes, which is 4,5,7,8, and use 256 filters. In total 1024 filters produce 1024 one-dimensional vectors and send them to the max-pooling layer to extract the feature with the largest activation degree. Splice all the features together, and finally pass through two linear fully-connected layers to get the final output. After 20 epochs, the loss is converging and the best accuracy is 63\%.

Third, We used DistilBERT(\cite{devlin2018bert}) for sentence meaning transform, and then built a 3 layer neural network to classify the sentence-level vector. For the DistilBERT pre-trained model, we apply the DistilBERT tokenizer on the sentence and then use the distilbert-base-uncased model to turn sentences into a vector. Each sentence is turned into a 768 dimensional vector. And then we build a three layer feedforward neural network with 512, 128, and 64 nodes to classify sentence vectors. The highest accuracy of this model is 57\%.

Lastly, we followed lukasgarbas’s BERT fineturing with ktrain \footnote{https://github.com/lukasgarbas/nlp-text-emotion}. In this fineturing, we did not directly use BERT as a sentence level embedding just like we did for DistilBERT.  The accuracy of this model is 83\% and this is our meme template selection model.

\subsection{Meme Caption Generator}
For both encoder and decoder, we used Pytorch to construct our model, followed by the structure in Picture \ref{Fig3} and Picture \ref{Fig4}. To train our model, we convert our sentences into tensors using the word2index dictionary, because we include the UNK token, we are able to handle unknown words.
	
To train our encoder model with the input sentence, we need to keep track of every output and the last hidden state for the decoder, and for the decoder “\textless SOS\textgreater” is the first input, the output and hidden state from the encoder are also taken to train the decoder.
	
During the training of the decoder, we use the previous target output as the input instead of the previous decoder output, this is called teacher-forced network. We did this because the model can converge faster. By using teacher forcing, our model can learn the “grammar” of the target output, and this is what we needed: we need a model that can convert a normal english text into a “meme-style” text. With teacher forcing, our model learned how to generate “meme-style” text.
	
For the evaluation of the model, we do not have a gold standard target output, so we use the previous output of itself, we store every output and stop when we have a special token “\textless EOS\textgreater”. We also record the loss during the training.

\section{Results \& Discussion}
\subsection{Meme Template Selection Model}
We use accuracy and f1 score to help select our final model, for all four models, we take the accuracy and f1 score and show them in Table \ref{Table2}.

\begin{table}[htbp]
	\centering
	\begin{center}
		\begin{tabular}{|c|c|c|}
			\hline Method & Accuracy & F1\\\hline
			LSTM & 61.27$\%$ & 61.95$\%$\\
            TextCNN & 62.86$\%$ & 63.45$\%$\\
            DistilBERT & 57.32$\%$ & 60.31$\%$\\
            BERT & 83.29$\%$ & 82.37$\%$\\
			\hline
		\end{tabular}
	\end{center}
	\caption{All models' performance}
	\label{Table2}
\end{table}

For BiLSTM and TextCNN we get pretty close results, but for DistilBERT we get an under performance model. Our discussion about the DistilBERT model is that: Because we only used BistilBERT as an embedding method, and we did not use a transformer based pre-trained classifier. Professor Ma suggested that we can finetune transformer based classifiers, so we did more research on the BERT and ktrain library. In the end we finetune and get a reasonably accurate BERT text classifier and decided to use it as our meme template selection model.

\subsection{Meme Caption Generator}
\begin{figure}[H] 
\centering
\includegraphics[width=0.4\textwidth]{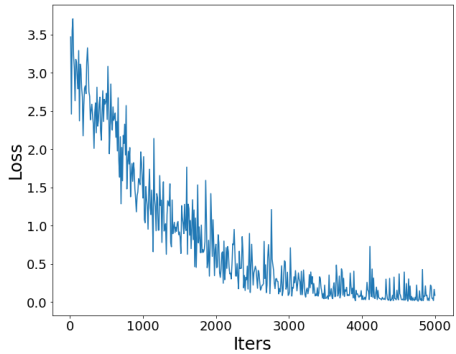} 
\caption{loss of the training process}
\label{Fig3}
\end{figure}

The training process of encoder and decoder meet our expectation when we keep tracking our loss, the loss of the training process is shown in Figure \ref{Fig3}.

During the evaluation process, we considered several questions. Is the caption understandable? Is the caption related to the input text? Is the caption suitable for a meme? During the evaluation, we found out that the model can do a good job when the test input is grammarly close to the train data, and when the input text is very grammarly different from the training data, the model will not do a good job like it did before.

We also found out that our model is good at generating “meme style” and context related output. One reason for this is that our training dataset is small, and a lot of words are reorganized as UNK. We have some examples of evaluations in Table \ref{Table3} and memes in Figure \ref{Fig4}.

\begin{figure}[htbp] 
\centering
\includegraphics[width=0.45\textwidth]{discussion8.pdf} 
\caption{Memes}
\label{Fig4}
\end{figure}

\begin{table}[H]
\footnotesize
	\centering
    	\begin{center}
    	\scalebox{0.7}{
    		\begin{tabular}{|c|c|c|}
    			\hline Input text & Output Caption & Evaluation\\\hline
    			\makecell{I need to do homework \\ before it due} & \makecell{I DO NOT ALWAYS DO \\HOMEWORK BUT WHEN I \\DO IT IS GOING TO DUE} & Good\\
    			\hline
    			\makecell{My cat shit on the floor, I am\\ so mad right now} & \makecell{I AM CAT AND SHIT \\FLOOR MAD} & Good\\
    			\hline
    			\makecell{i don't understand physics} & \makecell{WTF IS MY PHYSICS ! I\\ HATE IT SO MUCH} & Good\\
    			\hline
    			\makecell{I need money, i am going \\to rob a bank} & \makecell{NEED TO MONEY ? ROB A\\ BANK} & Acceptable\\
    			\hline
    			\makecell{i love NLP} & \makecell{I FEEL LOVE BECAUSE I \\AM NLP} & Acceptable\\
    			\hline
    			\makecell{Wanna hang out tonight?} & \makecell{FEEL WANNA BECAUSE\\ HANG OUT TONIGHT} & Bad\\
    			\hline
    		\end{tabular}}
    	\end{center}
	\caption{Examples of evaluations}
	\label{Table3}
\end{table}

In total we write 100 sentences for the test by separating similar and different sentences compared to the train sentences, we have 60 similar sentences and 40 different sentences, the evaluation report is in Table \ref{Table4}.

\begin{table}[htbp]
\footnotesize
	\centering
	\begin{center}
		\begin{tabular}{|c|c|c|c|}
			\hline & Good  & Acceptable & Bad\\\hline
			Similar & 47(78.33$\%$) & 8(13.33$\%$) & 5(8.33$\%$)\\
            Different & 15(37.5$\%$) & 18(45$\%$) & 7(17.5$\%$)\\
            total & 62(62$\%$) & 26(26$\%$) & 12(12$\%$)\\
			\hline
		\end{tabular}
	\end{center}
	\caption{Evaluation Metrics}
	\label{Table4}
\end{table}

In general our model can mostly produce a good or acceptable meme. But our evaluation is very subjective, and the size of both our training data, testing data are small. Within the limitation of dataset and evaluation method, we can still conclude that the encoder and decoder with attention structure is a possible approach for meme caption generation.

\section{Conclusions}
Generating memes is a challenging task, requiring complex image-text reasoning. In this project, we have successfully demonstrated how to use a neural model to generate memes and presented the Meme Generator, which is an end to end architecture that can automatically generate a meme for a given sentence. Meme Generator has two components, a model to select a meme template image and an encoder-decoder model to generate a meme caption. Both of the models are fine-tuned and achieved the current best performance.

We acknowledge that the biggest challenge in our project is choosing an appropriate model to train the data. We tried to use models defined by ourselves but the effect is very general. After we import the pre-trained model, it leads to a dramatic improvement of performance of the model. This also means that as the difficulty of the task increases, we cannot just use a simple network model.

There is another challenge in the project. For the emotion labels, they may vary from different cultures and people. In the future, we will take this element into consideration and want to cover more cultural backgrounds. We may introduce some new solutions from other papers, such as adding the break points in the text between upper and lower for the image.

For our model, it is still possible to be improved by different methods. One is to try to use a larger and more comprehensive dataset to let Meme Caption Generator generate caption better. The other is to introduce a better evaluation method. However, the quality of Memes varies among people. It is very difficult to evaluate them. A good meme is usually added and shared many times on social media. At this stage, we are just artificial evaluation and divide it into good, fair and bad. For future work we can evaluate memes by introducing it in a social media stream and using transmission rate as an evaluation metric.

Also we can use a larger data set to train a more powerful meme template selection model. So far the data set we are currently using is a small data set compiled by our team members. Although the model produces considerable results, there is still a lot of room for improvement. Using a large and balanced data set so that it can identify more types of emotions and make the selected image template more precise.

\clearpage

\bibliographystyle{ACL}
\end{document}